\documentclass[lettersize,journal]{IEEEtran}
\usepackage{amsmath,amsfonts}
\usepackage{algorithmic}
\usepackage{algorithm}
\usepackage{array}
\usepackage[caption=false,font=normalsize,labelfont=sf,textfont=sf]{subfig}
\usepackage{textcomp}
\usepackage{stfloats}
\usepackage{url}
\usepackage{verbatim}
\usepackage{graphicx}
\usepackage{cite}
\usepackage{color}
\usepackage{amsfonts}
\usepackage{balance}

\begin{document}

\title{End-to-end Speech Recognition with similar length speech and text}

\author{Peng Fan, Wenping Wang, Fei Deng

\thanks{\textit{Corresponding author: Fei Deng.}}

\thanks{Peng Fan and Fei Deng are with the College of Computer Science and Cyber Security, Chengdu University of Technology, Chengdu 610059, China.(e-mail: fanpeng2023@hotmail.com; dengfei@cdut.edu.cn).}
\thanks{Wenping Wang is with National Key Laboratory of Fundamental Science on Synthetic Vision, Sichuan University, Chengdu 610065, China. (email:sonography@aliyun.com)}

}

\markboth{Journal of \LaTeX\ Class Files, Vol. 14, No. 8, August 2015}
{Shell \MakeLowercase{\textit{et al.}}: Bare Demo of IEEEtran.cls for IEEE Journals}
\maketitle

\begin{abstract}

The mismatch of speech length and text length poses a challenge in automatic speech recognition (ASR). In previous research, various approaches have been employed to align text with speech, including the utilization of Connectionist Temporal Classification (CTC). In earlier work, a key frame mechanism (KFDS) was introduced, utilizing intermediate CTC outputs to guide downsampling and preserve keyframes, but traditional methods (CTC) failed to align speech and text appropriately when downsampling speech to a text-similar length. In this paper, we focus on speech recognition in those cases where the length of speech aligns closely with that of the corresponding text. To address this issue, we introduce two methods for alignment: a) Time Independence Loss (TIL) and b) Aligned Cross Entropy (AXE) Loss, which is based on edit distance. To enhance the information on keyframes, we incorporate frame fusion by applying weights and summing the keyframe with its context 2 frames. Experimental results on AISHELL-1 and AISHELL-2 dataset subsets show that the proposed methods outperform the previous work and achieve a reduction of at least 86\% in the number of frames.

\end{abstract}
\begin{IEEEkeywords}
speech recognition, end-to-end, self-attention, frame reduction
\end{IEEEkeywords}
\section{Introduction}
\label{sec:intro}

Recently, the Conformer-based end-to-end automatic speech recognition (ASR) model has been the most popular way in the speech recognition research community. According to the decoder's type, end-to-end ASR model can be divided into Connectionist Temporal Classification (CTC) based
model, attention-based Encoder-Decoder (AED) model, and  RNN-Transducer (RNN-T) based model \cite{2016Listen,2018Speech,graves2006connectionist,graves2012sequence,zhang2021tiny,graves2014towards}.

In the field of speech processing, the challenge of speech length exceeding text length is a noteworthy concern, particularly in tasks related to speech translation and speech recognition. To address this issue, several methods have been developed. For example, text length can be extended to match the length of the speech, as demonstrated in GMM-HMM. Also, alignment of speech and text can be achieved by introducing blank labels through CTC and RNN-T. Another approach is to integrate the speech features of the encoder into the encoder's autoregressive decoding process via AED, without the need for explicit alignment~\cite{2016Listen,graves2006connectionist,graves2012sequence,aymen2011hidden}. These solutions partially solve the problem of speech text mismatch, but the length of speech remains considerable, as does the computational workload. During speech recognition process, speech is usually downsampled by 3-4 times the length, but the speech length is still much longer than the text length. In addition, previous work used CTC outputs to guide downsampling and skip the blank frame for decoder~\cite{wang2023accelerating,yang2023blank,tian2021fsr}. 

Recently, some researchers employs an intermediate CTC ~\cite{interctc} to downsampling the speech feature in encoder, key frame-based downsampling (KFDS)~\cite{10296536} and Skipformer~\cite{10687583}, akin to the approach introduced by Wang et al.~\cite{wang2023accelerating} for accelerating the process, incorporating a CTC output guide for the downsampling task. Meng's work uses intermediate CTC and CTC Spike Reduction methods to guide attention masks and reduce redundant peaks, thereby increasing model efficiency~\cite{meng2024seq}.

In this study, our work is based on KFDS. We only retain key frames and downsample them to a length similar to the text. However, a new problem arises, that is, CTC loss cannot be applied if the speech length is similar to the text length. Therefore, we introduce the Length Similarity Loss (LSL) to address this issue. LSL comprises two implementation methods: one is the Time Independence Loss (TIL), which removes temporal information from both the input and output. The other is the Aligned Cross-Entropy (AXE) loss, which relies on edit distance alignment to synchronize the input and output before computing the cross-entropy loss. Our work is akin to CIF~\cite{dong2020cif} in terms of achieving alignment between input and output. However, a notable difference is that CIF does not incorporate speech downsampling.

In this work, we focus on ASR with the aim of downsampling speech to a length similar to the corresponding text. Experimental results on the AISHELL-1 dataset and AISHELL-2 dataset subsets demonstrate the effectiveness of our proposed methods, achieving a Character Error Rate (CER) comparable to the baseline. Furthermore, our approach performs comparably to previous methods that utilized CTC loss. Meantime, substantial reduction in computational complexity is achieved, while reducing the number of frames by at least 86\%.

\section{Method}

\label{sec:guidelines}
\begin{figure*}[t!]
\centerline{\includegraphics[width=0.95\linewidth]{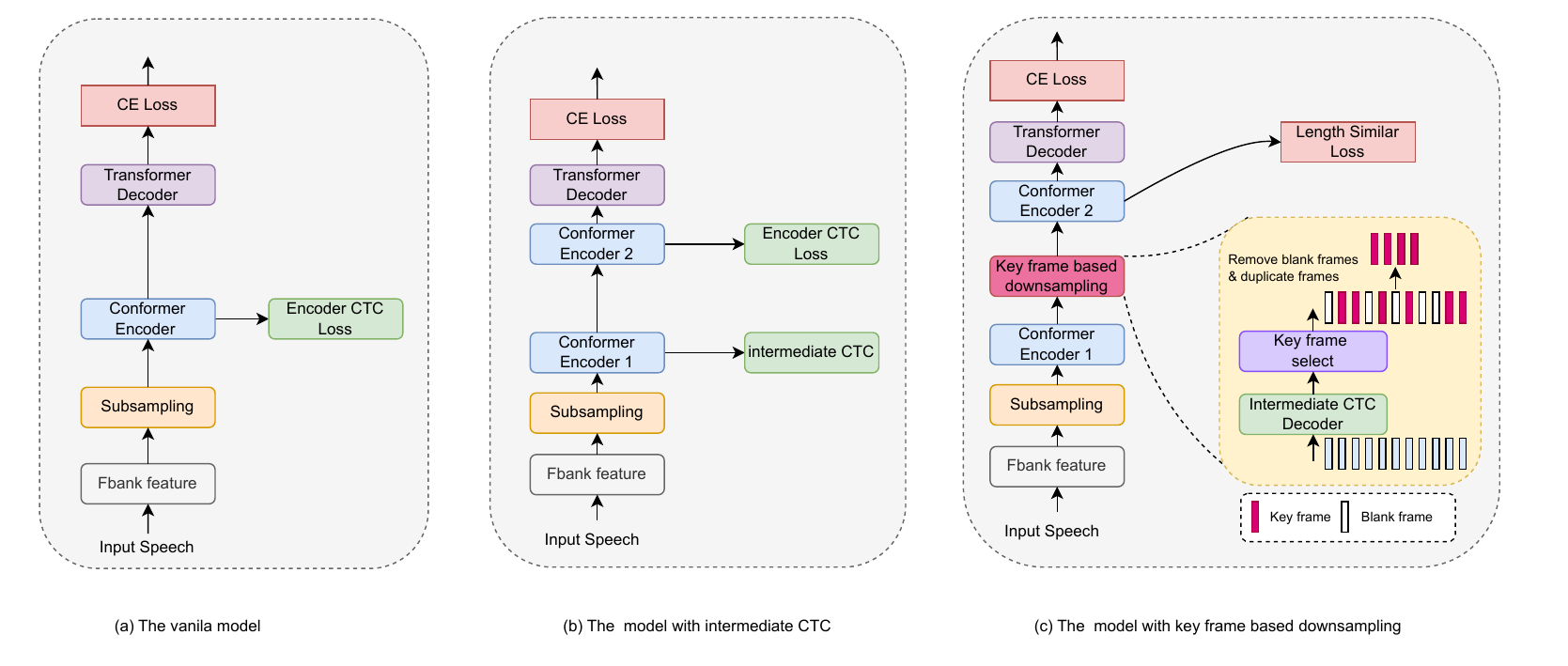}}
 	\caption{The overall architecture of the vanilla Conformer-based AED model (a), the AED model with intermediate CTC (b), and the proposed speech length is similar to the text model(KFDS-based mechanism downsampling) (c).}\label{fig1}
\end{figure*}

In this paper, our end-to-end ASR model comprises Conformer-based encoder layers and Transformer-based Decoder layers. Specifically, the vanilla AED joint CTC loss model, shown in Fig. 1(a), incorporates the CTC loss function during training to acquire speech-to-text alignment information, thereby enhancing the performance of the AED model. The final objective loss function is defined as follows:

\begin{equation}
\mathcal{L}=\alpha_0*\mathcal{L}_{ctc}+ \alpha_1 * \mathcal{L}_{ce}.
\end{equation}

Fig. 1(b) illustrates the AED model with intermediate CTC, which enhances model performance. The final objective loss for this model is defined as follows:

\begin{equation}
\mathcal{L}=\alpha_0*\mathcal{L}_{inter-ctc}+ \alpha_1 * \mathcal{L}_{CTC}+ \alpha_2 * \mathcal{L}_{ce}.
\end{equation}

As shown in Fig. 1, different from the vanilla conformer-based AED model, our proposed model incorporates a key frame-based downsampling module (the rose red block in Fig. 1(c)). Furthermore, the proposed LSL replaces CTC loss for computing the encoder's final output and target text. The proposed model's final objective loss is defined as:

\begin{equation}
\mathcal{L}=\alpha_0*\mathcal{L}_{inter-ctc}+ \alpha_1 * \mathcal{L}_{ls}+ \alpha_2 * \mathcal{L}_{ce}.
\end{equation}

Here, we set the parameters as follows: $\alpha_0, \alpha_1, \alpha_2 = 0.2, 0.1, 0.7$, employing the intermediate CTC loss to predict key frames and provide time information after encoder 1, LSL is uesd for handling the output after downsampling from encoder 2, and CE loss is utilized for the attention-based decoder output.

\subsection{Key frame downsampling}

We employ KFDS to downsample the length of speech to be similar to the text and subsequently explore speech recognition in this context. We will briefly introduce the key frame-based self-attention (KFSA) and the down-sampling process based on the key frame mechanism (KFDS). As shown in the dashed box at the bottom right of Fig. 1, the key frame mechanism selects key frame using the non-blank frame sequence generated by the intermediate CTC loss to remove duplicates and blank frames~\cite{wang2023accelerating,interctc}. 

The KFSA mechanism utilizes previously generated keyframes to reduce the self-attention mechanism module. The KFDS process, as illustrated in Fig. 1, involves down-sampling frames guided by key frames and preserving the frames corresponding to these key frames.

\subsection{Length Similar Loss (LSL)}

However, in prior research, the final output of the encoder was determined using the CTC loss. When we use the KFDS mechanism to downsample speech to a length similar to the text, the speech and text are mapped to the same feature space, and the previous method (CTC) is not suitable for calculating losses. 

This subsection, we introduce the LSL to deal with it. We introduce two LSL functions to address this issue from two distinct angles. Firstly, we employ a time information removal method (TIL) to compute the loss between the model's predictions and the ground truth values. Secondly, we align these elements by adjusting the distance before calculating the CE loss (AXE loss).

\subsubsection{Time independence loss (TIL)}


In this work, after frame-downsampling by the KFDS mechanism, the length of the output of the encoder is similar to the target text. However, their corresponding relationship cannot be established directly. The calculation of loss is prevented due to the absence of chronological order and one-to-one correspondence. Consequently, the temporal information was removed, and all information sets were consolidated into a single output denoted as $Y^{\prime}$.

Note that, despite the removal of timing information at the end, it has already been acquired by the inter-CTC model in the sixth layer, ensuring that the final output will not be out of order.

Let $Y$ be a target sequence of $L$ tokens $\{y_1, y_2, \ldots, y_L\}$, and $P$ be the model predictions, a sequence of $T$ token probability distributions $\{p_1, p_2, \ldots, p_T\}$, $v$ is the vocabulary size. We ignored the time information of the target $Y$ and predicted $P$. $Y^{onehot} = \{y^{onehot}_1, y^{onehot}_2, \dots, y^{onehot}_L\} $ correspond $Y$ one-hot vector of $v$ dimension. The time independence loss is defined by Eq.6.

\begin{equation}
     Y^{\prime} = \sum_{i=1}^{L}{y^{onehot}_i} 
\end{equation}

\begin{equation}
    P^{\prime} = \sum_{i=1}^{T}{p_i}
\end{equation}

\begin{equation}
    TIL = \sum Y^{\prime}\log{\frac{Y^{\prime}}{P^{\prime}} } 
\end{equation}

\subsubsection{Aligned cross-entropy loss (AXE)}
The AXE loss was initially proposed to solve the problem of excessive penalty for the change of output word order when using CE loss in machine translation\cite{ghazvininejad2020aligned}. The final prediction sequence of our encoder is very close to the target sequence in length, but there may be insertion and deletion errors, which is unsuitable for directly using CE loss, so the AXE loss function is introduced to calculate the loss.
Our goal is to find a monotonic alignment between $Y$ and $P$ that will minimize the CE loss, and thus focusing the penalty on lexical errors (predicting the wrong token) rather than positional errors (predicting the right token in the wrong place). We define an alignment $\alpha$ to be a function that maps target positions to prediction positions, i.e. $ \{1,\ldots, L\} \to \{1,\ldots, T\}$. The AXE loss can be depicted with the following Eq.7~\cite{ghazvininejad2020aligned}. 

\begin{equation}
    \begin{array}{l}
    \setlength{\arraycolsep}{3.0pt}
\operatorname{AXE}\left(Y_{1}, \ldots, Y_{L}, P_{1}, \ldots, P_{T}\right)= \\
\\
\min \limits _{\alpha} \operatorname{AXE}\left(Y_{1}, \ldots, Y_{L}, P_{1}, \ldots, P_{T} \mid \alpha\right)= \\ \\
\min \limits _{\alpha(1) \ldots \alpha(L)}\left(-\sum\limits _{i=1}^{L} \log P_{\alpha(i)}\left(Y_{i}\right)\right. \\
\left. -\sum \limits _{k \in\{1 \ldots T\} \backslash\{\alpha(1), \ldots \alpha(L)\}} \log P_{k}(\varepsilon) \right) \\
\\
\text { s.t. } 1 \leq \alpha(1) \leq \alpha(2) \leq \alpha(3) \ldots \leq \alpha(L) \leq T
\end{array}
\end{equation}

\subsection{Frame fusion}
In this paper, we use the keyframe mechanism for downsampling. Benefiting from the two proposed loss calculation methods, we can achieve extreme downsampling of speech similar in length to the text. If only keyframes are used, a lot of useful information will be discarded. Although CTC has peaks, its adjacent frames also contain a large amount of information. Therefore, we used frame fusion to preserve more information and employed two methods for frame fusion.

\begin{figure}[ht!]
\centerline{\includegraphics[width=0.55\linewidth]{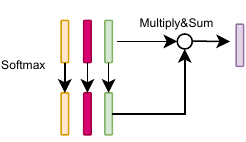}}
 	\caption{Attention-based frame fusion. }\label{fig1}
\end{figure}

Firstly, frame fusion was achieved through attention-based frame fusion without adding new parameters. As shown in Fig.2, the rose red bar represents the key frame, the orange bar represents the left frame of the keyframe, and the light green bar represents the right frame of the key frame. This process is shown in the following Eq. 8 and Eq. 9.

\begin{equation}
    \mathbf{S} = Softmax(h_{t-i : t+j})
\end{equation}

\begin{equation}
    O = \sum_{k = t-i}^{t+j} \mathbf{S}_{k}\cdot (h_{k})
\end{equation}

Here, $t$ is key frame, $\mathbf{S}$ represents the weight of frames from $t - i$ to $t + j$, with $i$ equal to 1 and $j$ equal to 1, while $O$ denotes the fusioned 3-frame.

As shown in Fig.3, we also used concatenate-based frame fusion, concatenated the keyframe sequence with the keyframe context 2 frames on the channel dimension, and then reduced it to the original dimension by using a linear layer.
\begin{figure}[ht!]
\centerline{\includegraphics[width=0.60\linewidth, ]{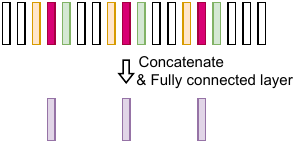}}
 	\caption{Concatenate-based frame fusion. }\label{fig2}
\end{figure}

\subsection{Three-step trainning}

\begin{figure}[ht!]
\centerline{\includegraphics[width=0.95\linewidth]{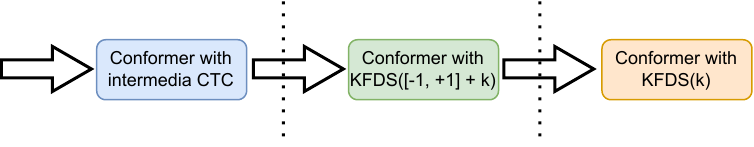}}
 	\caption{Three-step training. }\label{fig1}
\end{figure}

To facilitate model training convergence, a three-stage training process is proposed. The training process of our proposed model is depicted in Fig.4. In the first stage, a base Conformer model is trained to utilize intermediate CTC loss to obtain peak information. Subsequently, in the second stage, the KFDS method with key frame and context 2 frames ([-1,+1]+K), initializes the model using the Stage 1 model init for downsampling speech length. In the third stage, to achieve further downsampling, LS loss is adopted, and the model is initialized using the Stage 2 model.

\section{EXPERIMENTS}
\label{sec:guidelines}
\subsection{Corpus and  Experimental Configurations}

In this subsection, we verify our proposed KFDS-based Conformer encoder-decoder network on the open-source datasets: AISHELL-1~\cite{bu2017aishell} and AISHELL-2~\cite{du2018aishell} subsets. In all our experiments, 80-dimensional log Mel-filter bank (Fbank) features are extracted from a 25ms window with a 10ms frame shift. SpecAugment is used as acoustic feature augmentation~\cite{park19e_interspeech}. To conduct modeling on the AiSHELL-1, a vocabulary consisting of 5234 labels that incorporate Chinese characters and other special characters is employed.

We adopt hybrid CTC and attention-based auto encoder-decoder (AED) architecture following Wenet recipe \cite{zhang2022wenet}. For the encoder, there are 12 Conformer blocks. For each block, the convolutional kernel size is 31, the number of attention heads is 4, and the hidden dimensions of the attention and FFN layer are 256 and 2048 respectively. The decoder consists of 6 Transformer blocks, where each block has 4 attention heads.

As for training details, we follow the training recipes provided by Wenet. \footnote{The recipe on AISHELL-1 is publicly available in \url{https://github.com/wenet-e2e/wenet/tree/main/examples/aishell/s0/conf};} It will be easy for others to reproduce our experiments. Note that, in order to obtain a better initial intermediate CTC guidance, KFSA and KFDS are introduced after the first $N$ normal training epochs. $N$ is 40 for AISHELL-1 experiments.

\subsection{Overall Results}

\begin{table}[h]
\centering    

\caption{The overall result on AISHELL-1 with CER.}
\label{ai1}
\setlength{\tabcolsep}{1mm}
\begin{tabular}{llllc}
\hline
Models                   & Condition                                                                        & Test  & Drop ratio \\ \hline
Conformer\cite{zhang2022wenet}       & All                                                                          & 4.75 & /           \\
B0 (baseline)                     & All                                                                          & 4.58 & /           \\

Efficient Conformer\cite{burchi2021efficient}                      & All                                                                          & 4.64 & /           \\
\hline
E1 (KFSA)                             &  K         & 4.58 & /           \\
E2 (KFDS)                             & {[}-1, +1{]} + K    & 4.52 & 65\% \\
\hline 
E3 (KFDS with TIL)                             &  K         & 4.65 & 87\%           \\

E4 (KFDS with AXE)                             &  K         & \textbf{4.49} & \textbf{87}\%           \\
E5 (KFDS without Loss)                             &  K         & 89.83 & 87\%           \\

\hline

\hline
\end{tabular}
\end{table}

\subsubsection{Results On Aishell-1}

TABLE~\ref{ai1} shows the overall character error rate (CER) on the AISHELL-1 test set. During decoding, CTC prefix beam search is used to generate N-best candidates first and then rescored using a Transformer decoder. We only report the final results here after the Transformer decoder rescore. In Table I, the result of the vanilla Conformer model was from \cite{zhang2022wenet}. However, there is no intermediate CTC during model training in \cite{zhang2022wenet}. For a fair comparison, we add intermediate CTC loss during training of B0 model, which is our baseline model with 4.58\% CER. We also compare our methods with Efficient Conformer \cite{burchi2021efficient}, which downsampled feature sequences uniformly.

Model E1 is trained using the previous KFSA mechanism with “+ K”, which means all other key frames are used during attention calculation. E1 is trained using key frames only and obtains 4.58\% 9.CER, which also proves that key frames contain more helpful information and are crucial for attention mechanisms. E2 is trained using the KFDS mechanism with local temporal context widths 1 ([-1, +1] + K). Compared with the KFSA-based models E1, the KFDS-based models obtain lower, 4.52\% CER and less computational complexity. E1 and E2 are both the previous key frame mechanism models.

Models E3 and E4 are trained using the specified KFDS mechanism, which uses only keyframes to guide the downsampling of speech frames. Both E3 and E4 adopt extreme downsampling, which reduces the speech frame to the length of the deduplicated non-blank label sequence predicted by the intermediate CTC loss. The E3 and E4 obtain 4.65\% and 4.49\% CER respectively. E3's performance surpasses the vanilla Conformer and is comparable with B0 and Eﬀicient Conformer. However, it performed wose compared to E1 using the KFSA mechanism and E2 using 3 frames of KFDS. This is likely because this method ignores the order information of the text, which is important in speech recognition. However, this method is also a tradeoff one for calculating the loss when the length of a speech frame sequence is close to that of a text sequence. 

Especially, E4 obtained 4.49\% CER, which surpasses the previous best results with 0.03\% absolute CER. This is a very interesting result. When using the KFDS mechanism, E4 with less information actually performs better than E2, indicating that the key frame mechanism has indeed learned enough key information. Meanwhile, compared to E3, the absolute CER decreased by 0.16\%, indicating that the time information in the speech frame is very important for predicting the final text sequence. E5 employed an extreme downsampling strategy without computing the loss for the output of the second encoder. Consequently, it achieved a CER of 89.83\%, leading to performance degradation. This underscores the crucial necessity of calculating the loss for the encoder 2 output. Additionally, our approach achieves an 87\% reduction in frames on the AISHELL-1 dataset, surpassing the previous reduction of 65\%.

It is worth mentioning that considering that the model is not easy to converge after extreme downsampling, we use E2's final average of 30 epochs model as model init to facilitate better training of our model.

\begin{table}[h]
\centering    

\caption{The overall result on AISHELL-2 subsets with CER.}
\label{ai2}

\setlength{\tabcolsep}{1mm}
\begin{tabular}{llllc}
\hline
Models                   & Condition                                                                        & Test  & Drop ratio \\ \hline
Conformer\cite{zhang2022wenet}       & All                                                                          & 8.40 & /           \\
B0 (baseline)                     & All                                                                          & 8.30 & /           \\

Efficient Conformer\cite{burchi2021efficient}                      & All                                                                          & 8.42 & /           \\
\hline
E1 (KFSA)                             &  K         & 8.21 & /           \\
E2 (KFDS)                             & {[}-1, +1{]} + K    & \textbf{8.18} & 58\% \\
\hline 
E3 (KFDS with TIL)                             &  K         & 8.76 & 86\%           \\

E4 (KFDS with AXE)                             &  K         & 8.43 & \textbf{86}\%           \\
E5 (KFDS without Loss)                             &  K         & 87.52 & 86\%           \\

\hline

\hline
\end{tabular}
\end{table}
\subsubsection{Results On Aishell-2 subsets}

To verify the effectiveness of our proposed method, we conducted experiments on subsets of the AISHELL-2 dataset. TABLE~\ref{ai2} presents the overall CER on the AISHELL-2 test set. The baseline B0 model with intermediate CTC loss achieves a CER of 8.30\%, while the vanilla Conformer has a CER of 8.40\%, demonstrating that the intermediate CTC loss method enhances model performance. In addition, the Efficiency Conformer model achieves a CER of 8.42\%, performing slightly worse than the first two baseline models but with reduced computational complexity. The KFSA model reaches a CER of 8.21\%, and the KFDS model achieves 8.18\%, demonstrating the effectiveness of the keyframe mechanism. For the proposed method in this study, TIL yields a CER of 8.76\%, the lowest performance among all models, attributed to information loss; temporal information is removed after feature downsampling. The AXE loss model, with a CER of 8.43\%, performs worse than both the B0 baseline model and the Efficiency Conformer but achieves substantial downsampling by discarding 86\% of the frames. Lastly, when the loss is not computed after the second encoder that performs downsampling, the model's CER increases significantly to 87.52\%, indicating a substantial degradation in performance. This aligns with the results observed on the AISHELL-1 dataset, underscoring the importance of calculating the loss at this stage.


\subsubsection{Frame fusion}

We first explored whether frame fusion is necessary for our experiment and which fusion method is optimal. Note that all models in this experiment were trained through pseudo-one-hot with KL loss. As shown in TABLE~\ref{fusion_lab}, E6 is a KFDS that only uses keyframes without any frame fusion, but it achieved the worst result \textcolor{black}{4.81\%}, which is worse than the E2 of the original KFDS mechanism and even worse than the performance of the three baseline models. By fusing 3 frames into 1 frame, E7 achieved a poor CER effect of 4.71\% using an additional network for frame fusion, compared to 4.65\% using weighted summation for E8. Finally, we attempted to fuse 5 frames into 1 frame and obtained a CER of 4.74\% using a weighted sum method, indicating that the effect of fusing more frames would also degrade, possibly due to the inclusion of information unrelated to the current keyframe.

\begin{table}[h]
    \centering
    \caption{
Investigate the different frame fusion methods}
    \label{fusion_lab}
    \begin{tabular}{ccc}
    \hline
Models   &Fusion method &Test   \\
    \hline
E6      &  No fusion & 4.81         \\
E7      &  Concatenate-based & 4.71          \\
E8      &  Attention-based(3 frames) & \textbf{4.65}           \\
E9     &  Attention-based(5 frames) & 4.74           \\
\hline
    \end{tabular}
   
\end{table}

\section{conclusion}

Matching the length of a speech with the length of its target text presents a significant challenge in the field of ASR. In this paper, we investigate ASR within the context of downsampling speech to match the length of the text using KFDS. In response to this challenge, we introduce two novel alignment techniques: a)TIL and b) AXE Loss. To enhance the content of key frames, we implement a frame fusion approach, incorporating weights and summing the key frame with its contextual two frames. Our proposed methodology has exhibited results that superior to the baseline in extensive experiments conducted on the AISHELL-1 dataset. Furthermore, our approach achieves a substantial reduction of at least 87\% in the number of frames. Further experiments on a subset of AISHELL-2 reinforce the effectiveness and robustness of our proposed method.


\vfill\pagebreak

\bibliographystyle{IEEEtran}
\bibliography{refs}

\end{document}